\documentclass[11pt]{article}

\usepackage[utf8]{inputenc}
\usepackage[T1]{fontenc}
\usepackage{amsmath}
\usepackage{amssymb}
\usepackage{geometry}
\usepackage{algorithm}
\usepackage{algpseudocode}
\usepackage{xcolor}
\usepackage{listings}
\usepackage{graphicx}
\usepackage{hyperref}
\usepackage{cite}
\usepackage{lmodern}

\title{The Free Will Equation: Quantum Field Analogies for AGI}
\author{Rahul Kabali}
\date{July 04, 2025}

\begin{document}

\maketitle

\begin{abstract}
Artificial General Intelligence (AGI) research traditionally focuses on algorithms that optimize for specific goals under deterministic rules. Yet, human-like intelligence exhibits adaptive spontaneity---an ability to make unexpected choices or free decisions not strictly dictated by past data or immediate reward. This trait, often dubbed ``free will'' in a loose sense, might be crucial for creativity, robust adaptation, and avoiding ruts in problem-solving. This paper proposes a theoretical framework, termed the Free Will Equation, that draws analogies from quantum field theory to endow AGI agents with a form of adaptive, controlled stochasticity in their decision-making process. The core idea is to treat an AI agent's cognitive state as a superposition of potential actions or thoughts, which collapses probabilistically into a concrete action when a decision is made---much like a quantum wavefunction collapsing upon measurement~\cite{ref1}. By incorporating mechanisms analogous to quantum fields, along with intrinsic motivation terms, we aim to improve an agent's ability to explore novel strategies and adapt to unforeseen changes. Experiments in a non-stationary multi-armed bandit environment demonstrate that agents using this framework achieve higher rewards and policy diversity compared to baseline methods.
\end{abstract}

\section{Introduction}
Artificial General Intelligence (AGI) research traditionally focuses on algorithms that optimize for specific goals under deterministic rules. Yet, human-like intelligence exhibits adaptive spontaneity---an ability to make unexpected choices or free decisions not strictly dictated by past data or immediate reward. This trait, often dubbed ``free will'' in a loose sense, might be crucial for creativity, robust adaptation, and avoiding ruts in problem-solving. In classical AI systems (e.g., deep neural networks or reinforcement learners), behavior is largely determined by minimizing loss or maximizing reward, having little room for intrinsic randomness beyond what is injected via fixed exploration heuristics. This paper proposes a theoretical framework, termed the Free Will Equation, that draws analogies from quantum field theory to endow AGI agents with a form of adaptive, controlled stochasticity in their decision-making process. The core idea is to treat an AI agent's cognitive state as a superposition of potential actions or thoughts, which collapses probabilistically into a concrete action when a decision is made---much like a quantum wavefunction collapsing upon measurement~\cite{ref1}. By incorporating mechanisms analogous to quantum fields (with continuous evolution of a distributed state and sudden collapse events), along with intrinsic motivation terms, we aim to improve an agent's ability to explore novel strategies and adapt to unforeseen changes.

Our contributions include:
\begin{itemize}
    \item A quantum-inspired cognitive model for AI: the agent's mind is viewed as a multidimensional field of possible action-states, which can exist in superposition and then resolve to a single action (decision) upon interaction with the environment~\cite{ref1}. This formalism captures non-deterministic choice in a principled way.
    \item The Free Will Equation, which mathematically encodes the balance between goal-directed exploitation and exploratory ``free'' choice. It integrates a standard reward-driven policy with an intrinsic entropy/novelty drive analogous to quantum uncertainty.
    \item Comparison with existing AI architectures, highlighting parallels with reinforcement learning exploration (e.g., $\epsilon$-greedy, Boltzmann policies~\cite{ref3}), transformer generative models' use of temperature for randomness~\cite{ref8}, and evolutionary algorithms that utilize diversity or novelty search~\cite{ref5}. We show that while these methods include stochastic elements, the Free Will Equation offers a more adaptive, context-sensitive modulation of randomness---more akin to a conscious decision to explore.
    \item Example experiments with metrics such as adaptation curves and policy entropy (diversity) to illustrate the benefits. In a non-stationary environment, an agent empowered by the Free Will Equation adjusts its ``temperature'' on the fly and discovers new optimal behaviors faster than a conventional agent. All experimental code is included to encourage researchers to run, reproduce, and build upon these ideas.
\end{itemize}

We believe this interdisciplinary framework---bridging concepts from quantum physics, cognitive science, and machine learning---offers a novel perspective on designing AGI systems that are both robustly goal-seeking and intrinsically innovative. In the following, we detail the theoretical underpinnings (Section~\ref{sec:background}), formulate the Free Will Equation (Section~\ref{sec:free_will}), draw connections with existing AI methods (Section~\ref{sec:comparison}), demonstrate its potential via experiments (Section~\ref{sec:experiments}), and discuss implications and future directions (Section~\ref{sec:discussion}).

\section{Background: Quantum Uncertainty and Adaptive Decision-Making}
\label{sec:background}
\subsection{Quantum Analogy for Cognitive Uncertainty}
In quantum physics, a system's state is described by a state vector (or wavefunction) in a Hilbert space, which encodes probabilities for all possible outcomes. The system evolves smoothly and predictably according to the Schrödinger equation when not observed (unitary evolution), but when a measurement occurs, the state reduces or collapses randomly to a definite outcome~\cite{ref9}. Crucially, the exact result of a quantum measurement is indeterminate---only the probability distribution of outcomes can be known. This inherent uncertainty has often been invoked in discussions of free will: if the physical world is not strictly deterministic at the fundamental level, perhaps conscious agents could harness this indeterminism for free decision-making~\cite{ref7}. While the role of quantum effects in human brains is controversial, the analogy is conceptually appealing. We take inspiration from this dual-mode behavior: continuous evolution of a distributed mental state punctuated by stochastic choice events. In our context, an AGI's internal cognitive dynamics between decisions will be viewed as analogous to the deterministic evolution of a quantum state, and the final decision (action selection) as analogous to a measurement yielding a probabilistic outcome (collapse). We propose that thoughts in an AGI can be represented in a superposed form, exploring multiple possibilities at once, until the moment of choice forces a singular action to materialize~\cite{ref1,ref9}.

\subsection{The $\Psi$-Field of Potential Actions}
Pushing the analogy further, one can think in terms of quantum fields rather than just quantum states. In quantum field theory (QFT), particles (outcomes) are excitations of underlying fields permeating space. By analogy, we envision an AGI's decision space as a cognitive field spanning all possible actions or strategies. At any given state, the agent has a field of amplitudes or ``propensities'' $\Psi(s)$ for each possible action $a \in \mathcal{A}$. This $\Psi$-field evolves under the influence of external inputs (observations, rewards) and internal dynamics (goals, drives), somewhat like a field governed by a Lagrangian that includes interaction terms. Before a decision, the agent's mind is in a superposition of different action tendencies, each with some amplitude (similar to how a photon's electromagnetic field can be in a superposition of states). When the agent must act (analogous to an observation happening), this field ``collapses''---one particular action is realized corresponding to one mode of the field getting excited (and others suppressed). The probability of each action being chosen is given by the squared amplitude of its mode (echoing Born's rule in quantum mechanics)~\cite{ref9}. Thus, the Free Will Equation we seek will in part describe how these amplitudes are computed and how one outcome is sampled. Notably, this collapsible-field view emphasizes that multiple options coexist in the agent's decision process until the last moment, providing a formal way to think about indecision, exploration, and creativity within a single framework.

It’s important to clarify that our approach is analogical: we are not necessarily claiming the agent is performing quantum computing or that true quantum processes are involved. Rather, we borrow mathematical and conceptual tools from quantum theory to enrich AI decision models. This approach aligns with emerging ideas in theoretical AI. For instance, Moitra \& Banerjee (2025) propose a multidimensional cognitive state for human-like intelligence, where ``thoughts exist in a state of superposition and collapse into definite outcomes through conscious focus---analogous to quantum measurement''~\cite{ref1}. Similarly, cognitive scientists have explored quantum cognition models in which human decisions are better explained by quantum-like probability laws than by classical logic. Certain paradoxical decision behaviors (such as violating the sure-thing principle) can be modeled by assuming that a person’s belief state is a quantum superposition that doesn’t collapse until a decision is made, introducing interference effects. Haven \& Khrennikov (2013) note that an individual’s final choice can be unpredictable and that ``people’s beliefs interact, or become entangled with their eventual action'' in ways classical probability cannot capture~\cite{ref9,ref10}.

\subsection{Free Will in AI and Inductive Bias}
Why introduce a ``free will'' mechanism at all? One reason is robust adaptation. A completely greedy, deterministic policy might do well in a static, known environment but fail when conditions change or when encountering novel situations outside its training distribution. By contrast, biological agents (including humans) display exploratory behavior and seemingly random trial-and-error, which often helps in discovering new solutions or adapting to changes. This has echoes in AI as well: reinforcement learning (RL) practitioners have long noted that some form of exploration is necessary to avoid suboptimal convergence. Classic algorithms use tricks like $\epsilon$-greedy (take a random action with probability $\epsilon$) or Boltzmann/softmax action selection (sample actions with probabilities proportional to $\exp(Q(a)/T)$ for some temperature $T$)~\cite{ref3}. These are essentially hard-coded ``free will'' injections---the agent deviates from pure reward-maximization occasionally or in a graded fashion. Our work seeks to endogenize this exploration drive: instead of a fixed $\epsilon$ or temperature schedule, the agent itself (via the Free Will Equation) decides when and how strongly to explore based on its internal state and experience, analogous to a conscious agent deciding to ``try something completely different'' when routine fails.

Another reason is fostering creativity and open-ended search. In evolutionary algorithms, there’s a concept of novelty search, where instead of optimizing a fixed objective, the algorithm rewards agents for doing something new that hasn’t been seen before~\cite{ref5}. Lehman \& Stanley (2011) famously showed that novelty-driven evolution can outperform objective-driven evolution on tasks with deceptive local optima. The novelty metric ``creates a constant pressure to do something new'' by measuring how unique an individual’s behavior is~\cite{ref5}. This is reminiscent of how scientific or cultural evolution progresses---not by single-mindedly optimizing one metric, but by exploring diverse possibilities, many of which serendipitously lead to better solutions. An AGI endowed with an intrinsic free-will/novelty drive might avoid the narrow convergent behavior that plagues many AI systems (e.g., mode collapse in GANs or repetitive outputs in language models). In our analogy, we can think of the agent’s policy as having its own ``Hamiltonian'' composed of two parts: one part encodes the external objective (environment rewards) and the other encodes an intrinsic drive toward entropy or novelty. When the intrinsic part dominates, the policy state spreads out (superposition broadens); when the external objective dominates, the state focuses around known good actions. Managing this balance is the crux of the Free Will Equation.

\subsection{Summary}
By drawing from quantum mechanics, we gain a language to describe non-deterministic choice in a rigorous way (using probabilities and superposition), and by drawing from cognitive science and evolutionary theory, we emphasize the utility of such nondeterminism for intelligence. In the next section, we formulate the Free Will Equation that synthesizes these ideas into a concrete decision rule for an AI agent. We will see that it can be viewed as a generalization of existing stochastic policies (like softmax with adaptive temperature), with the novel element that the ``temperature'' or randomness is not fixed but dynamically modulated akin to collapse events triggered by the agent’s context.

\section{The Free Will Equation Framework}
\label{sec:free_will}
\subsection{Formalizing Superposition and Collapse}
Consider an agent in state $s$ (which could encompass its observations, memory, etc.). Let $\Psi(s)$ denote the agent’s cognitive state vector over its action space $\mathcal{A}$. We can write $\Psi(s)$ as a weighted sum over basis vectors corresponding to each possible action:
\[
\Psi(s) = \sum_{a \in \mathcal{A}} \psi_a(s) |a\rangle,
\]
where $|a\rangle$ is a basis state representing ``taking action $a$'' and $\psi_a(s)$ is the real propensity weight for action $a$. In a purely classical policy, the agent would deterministically pick the $a$ with maximal weight. Instead, by analogy to quantum theory, we interpret $|\Psi(s)|^2$ as giving the probability of selecting an action. The probability of selecting action $a$ is given by a softmax distribution:
\[
p(a|s) = \frac{\exp \left( \frac{Q(s,a) + \alpha I(s,a)}{T} \right)}{\sum_{a' \in \mathcal{A}} \exp \left( \frac{Q(s,a') + \alpha I(s,a')}{T} \right)},
\]
where $Q(s, a)$ is the expected reward, $I(s, a)$ is an intrinsic motivation term, $\alpha \geq 0$ scales its influence, and $T > 0$ is the temperature controlling exploration. The overall entropy of the policy (higher $T$ leads to more uniform probabilities, lower $T$ is greedier). Crucially, $I(s, a)$ is an intrinsic incentive term capturing the novelty, surprise, or uncertainty associated with taking action $a$ in state $s$. $\alpha$ is a scaling factor for how strongly intrinsic motivation influences the decision. This equation can be seen as one realization of the operator $H_{\text{int}}$, effectively contributing an extra ``energy'' $\alpha I(s, a)$ that makes uncommon or curious actions more likely than their external reward alone would justify.

Several choices of $I(s, a)$ are possible: for example, one might define $I(s, a)$ to be inversely proportional to the frequency with which action $a$ has been taken in similar states (encouraging novel actions), or related to the prediction error or surprise if $a$ were taken (encouraging information gain). In the simplest case, we define:
\[
I(s, a) = \frac{1}{\sqrt{1 + N(s,a)}},
\]
where $N(s, a)$ is the number of times action $a$ has been taken in state $s$. This encourages novel actions by assigning higher bonuses to less-tried options, inspired by count-based exploration~\cite{ref8}.

The field evolves between decisions via:
\[
\Psi_{t+1}(s) = F(\Psi_t(s), I_t, \theta_t),
\]
where $F$ is an update operator, $I_t$ represents external inputs (e.g., rewards), and $\theta_t = \{T_t, \alpha_t\}$ are adaptive parameters. Specifically:
\[
\psi_{a,t+1}(s) = \psi_{a,t}(s) + \eta \left[ r_t \cdot I(a_t = a) + \alpha_t I(s, a) - \psi_{a,t}(s) \right],
\]
where $\eta$ is a learning rate, $r_t$ is the reward, and $I$ is the indicator function. The temperature $T_t$ adapts based on surprise:
\[
T_{t+1} =
\begin{cases} 
\min(T_{\text{max}}, T_t \cdot \gamma_{\text{inc}}) & \text{if } |r_t - \bar{r}_t| > \tau, \\
\max(T_{\text{min}}, T_t \cdot \gamma_{\text{dec}}) & \text{otherwise},
\end{cases}
\]
where $\bar{r}_t$ is the moving average reward, $\tau$ is a surprise threshold, and $\gamma_{\text{inc}}, \gamma_{\text{dec}}$ are increase/decay factors.

\subsection{Implementation Perspective}
From an implementation point of view, the Free Will Equation can be integrated into an algorithm as follows. We maintain usual value estimates (or a learned policy network) for external rewards. Additionally, we track metrics for novelty or uncertainty---for example, a count of how many times each action has been tried in a given context, or the variance of recent outcomes, or the error in the agent’s predictive model for that state. At decision time, we combine these to compute a final selection probability.

\begin{algorithm}
\caption{Free Will Action Selection}
\begin{algorithmic}[1]
\State Initialize $\epsilon = 0.5$, $T_{\text{min}} = 0.01$, $T_{\text{max}} = 2.0$, $\gamma_{\text{inc}} = 1.05$, $\gamma_{\text{dec}} = 0.85$, $\tau = 0.4$
\For{each time step $t$}
    \If{surprise level $> \tau$} \Comment{Significant drop in recent reward}
        \State $T_t \gets \min(T_{\text{max}}, T_t \cdot \gamma_{\text{inc}})$
    \Else
        \State $T_t \gets \max(T_{\text{min}}, T_t \cdot \gamma_{\text{dec}})$
    \EndIf
    \For{each action $a$}
        \State Compute score: $Q(s,a)/T_t + \alpha \cdot \text{IntrinsicBonus}(s,a)$
        \State \Comment{IntrinsicBonus, e.g., $1/\sqrt{1 + \text{visit\_count}(s,a)}$}
    \EndFor
    \State Compute probabilities: $p(a|s) = \text{softmax}(\text{score})$
    \State Select action $a_t \sim p(a|s)$
    \State Observe reward $r_t$ and next state $s_{\text{next}}$
    \State Update $\psi_{a,t+1}(s) = \psi_{a,t}(s) + \eta \left[ r_t \cdot I(a_t = a) + \alpha_t I(s, a) - \psi_{a,t}(s) \right]$
    \State Update $Q(s, a_t) \gets Q(s, a_t) + \eta \left( r_t + 0.9 \cdot \max_a Q(s_{\text{next}}, a) - Q(s, a_t) \right)$
    \State Update $N(s, a_t) \gets N(s, a_t) + 1$
\EndFor
\end{algorithmic}
\end{algorithm}

In words, the agent monitors its performance and triggers a ``free will boost'' to exploration when needed. The variable \texttt{surprise\_level} could be a measure of how unexpected or disappointing recent outcomes have been---essentially detecting when the agent’s current policy is likely failing (this is analogous to a quantum measurement disturbing the system, prompting a collapse to a new state). When surprise is high, the agent increases its temperature (allowing more random deviations). Otherwise, it slowly decays the temperature to focus more on exploitation. The combination of $Q$ and intrinsic bonus in \texttt{score(a)} ensures even when temperature is moderate, the agent has a bias toward trying actions that are informationally promising (not just reward-promising). This approach encourages directed exploration---not pure randomness, but exploration with purpose.

To ground this in a metaphor: imagine an AGI robot scientist working in a lab. Most of the time, it follows the gradient of knowledge---doing experiments likely to yield high reward (new discoveries). But occasionally, when progress stalls (low reward change, high surprise), it deliberately stirs in some randomness---trying a wacky experiment or exploring a neglected avenue. It keeps track of what it has tried before, so it leans toward experiments that are novel (the intrinsic bonus) to maximize the chance of learning something new. This deliberate yet stochastic strategy is what the Free Will Equation formalizes.

\subsection{Relation to Quantum Mechanics Concepts}
It is worth explicitly mapping components of our framework to quantum concepts, reinforcing the analogy:
\begin{itemize}
    \item \textbf{Hilbert space}: The space of an agent’s possible mental states (spanned by basis $|a\rangle$ for each action or more complex basis including thoughts) is analogous to a Hilbert space. The agent’s state $\Psi$ is a vector in this space. High-dimensional cognitive representations can be seen through this lens~\cite{ref1}.
    \item \textbf{State superposition}: The agent maintains a distribution over actions rather than a single chosen action until the moment of decision (like a wavefunction spread over multiple eigenstates). For instance, in a planning phase, the agent might entertain multiple potential moves concurrently.
    \item \textbf{Unitary evolution vs. measurement}: The two kinds of state evolution in quantum theory correspond to (a) the agent updating its beliefs/values (a continuous, quasi-deterministic process, e.g., via gradient descent or Bayesian update), and (b) the agent committing to an action (a probabilistic jump). The lack of a known trigger for quantum collapse (the measurement problem~\cite{ref9}) is mirrored in our agent by a heuristic trigger for exploration---effectively, the agent itself decides when to ``collapse differently'' by raising its randomness, analogous to an observer effect.
    \item \textbf{Entanglement}: In quantum physics, entangled particles share linked states. In our context, entanglement could be a metaphor for how beliefs and actions become intertwined~\cite{ref9}. For example, an agent’s prior belief influences its action, which then influences its updated belief. We ensure through intrinsic terms that the agent doesn’t overconfidently tie its belief to a single action prematurely (avoiding a classical collapse too early).
    \item \textbf{Many-worlds vs. selection}: One interpretation of quantum mechanics (Many-Worlds) says all possible outcomes happen in branching universes, avoiding collapse. Similarly, one could envision an ensemble of agent instances exploring all possibilities in parallel (like a population in evolutionary algorithms). Our approach instead uses collapse (one world actualized) but ensures that over repeated trials, different possibilities will materialize, effectively sampling the space over time.
\end{itemize}

The Free Will Equation is thus not a single equation in the traditional sense, but a framework combining a stochastic policy distribution with an adaptive rule for randomness influenced by intrinsic drives. In the next section, we examine how this framework compares and connects to mainstream AI approaches.

\section{Comparison with Existing AI Paradigms}
\label{sec:comparison}
Our Free Will Equation approach intersects with several strands of AI research. Here we draw explicit comparisons to highlight similarities and differences:

\subsection{Reinforcement Learning (RL)}
In standard RL, exploration is often achieved by simple randomization strategies. For example, $\epsilon$-greedy policies choose a random action with probability $\epsilon$, otherwise the best-known action~\cite{ref3}. This is a crude form of free will injection---the agent has a fixed chance to act out of character. Similarly, Boltzmann/softmax exploration uses a temperature parameter to sample actions probabilistically, where lower temperature means more greedy (exploitative) and higher means more random~\cite{ref3}. These methods are usually hand-tuned or scheduled (e.g., annealing $\epsilon$ over time). By contrast, the Free Will Equation framework makes exploration state-aware and feedback-driven. Instead of a fixed $\epsilon$, the agent modulates its ``temperature'' based on surprise or uncertainty. This can be seen as a form of meta-learning the exploration rate. There is indeed recent interest in learning to explore in RL, and our approach provides a principled rationale for it (inspired by how humans appear to strategically explore). Moreover, intrinsic reward research in RL aligns well: adding an entropy bonus $H(p(a|s))$ to the loss encourages the policy to remain stochastic~\cite{ref3}, and using intrinsic rewards for novelty or curiosity encourages visiting unseen states~\cite{ref11}. Our intrinsic term $I(s, a)$ and adaptive temperature serve the same purpose, but packaged in the narrative of ``free will.'' Conceptually, one might say classical RL exploration is like ad-hoc quantum noise, whereas our approach aspires to be more like quantum-inspired self-determination---the agent chooses to be uncertain when needed, rather than uncertainty being merely a fixed perturbation.

\subsection{Large Language Models (Transformers)}
Modern generative models (like GPT series) produce outputs by sampling from a probability distribution of tokens. Users can control a parameter called temperature during sampling to adjust randomness: low temperature induces behavior close to greedy (most likely output), whereas high temperature leads to more uniformly random sampling~\cite{ref8}. This is directly analogous to our concept of raising or lowering the policy entropy. In fact, the term ``temperature'' itself is borrowed from statistical physics (Boltzmann distribution)~\cite{ref12}. A noteworthy point is that high temperature in LLMs can yield more creative and diverse outputs at the cost of coherence, while low temperature yields deterministic and precise outputs~\cite{ref3}. This trade-off is essentially what any intelligent agent faces: explore (be creative) versus exploit (be correct). Currently, LLMs rely on the user to set temperature globally for a task. One could imagine an LLM with a Free Will Equation internally: it might increase its own ``temperature'' when faced with an ambiguous or novel prompt (to brainstorm widely) and decrease temperature when the context demands precision (to give a focused answer). Such dynamic control could improve performance on complex tasks that require a mix of creativity and accuracy. Additionally, stochastic decoding strategies (like nucleus or top-$k$ sampling) further shape the randomness in transformers~\cite{ref8}. These can be seen as different ways of collapsing the superposition of next-token possibilities. Our framework doesn’t directly implement top-$k$ sampling, but the principle of biased but non-deterministic choice is the same. The Free Will Equation could inspire new decoding schemes where the model’s uncertainty triggers, say, a broader exploration of possible continuations (like momentarily increasing $k$ or the nucleus threshold if it becomes ``unsure'' how to continue a story).

\subsection{Evolutionary Algorithms and Novelty Search}
Evolutionary approaches maintain a population of solutions and rely on variation (mutation, crossover) plus selection. The presence of multiple individuals implicitly explores many possibilities in parallel---akin to the Many-Worlds idea where each individual is exploring a different branch. The Free Will Equation in a single agent can be thought of as collapsing many potential ``parallel selves'' into one action at a time, but across multiple decisions it tries out diverse actions that a population might have tried in parallel. Techniques like novelty search explicitly reward divergence from prior behaviors~\cite{ref5}, which is analogous to our intrinsic term. In fact, our intrinsic reward $I(s, a)$ could be defined as a novelty score (e.g., negative similarity to past visited state-action pairs), thereby pushing the agent towards behaviors an evolutionary algorithm might discover over generations. The difference is one of granularity: evolutionary algorithms explore at the population level across generations, whereas our agent explores within its own lifetime via stochastic choice. There is also a parallel in terms of regime shifts in evolution: once in a while, a radical mutation can change an individual drastically (perhaps analogous to a quantum leap). In our agent, a surge in exploration (high temperature) is like a radical mutation in policy---it may temporarily try very different actions, providing a chance to break out of local optima. Thus, the Free Will Equation can be seen as bringing the spirit of open-ended, divergent search into an online learning agent. It’s notable that some researchers view open-ended innovation in evolution as partly driven by random drift and neutral mutations (i.e., not solely by fitness pressure~\cite{ref5}). Likewise, our agent’s behavior is not 100\% driven by reward maximization; a portion of its ``decision Hamiltonian'' is effectively random or novelty-driven, which can lead to serendipitous discoveries that pure reward optimization would overlook~\cite{ref5}.

\subsection{Complexity Science and Self-Organization}
From a complexity theory perspective, systems that balance exploration and exploitation often reside at the edge of chaos, a regime conjectured to be optimal for adaptability. Too much order (determinism) and the system is rigid; too much chaos (randomness) and the system is unglued. The Free Will Equation is an attempt to keep an agent in that critical balance by endogenously regulating its ``chaos level'' (entropy). One might draw an analogy to simulated annealing, where a high temperature helps explore the search space and a slow cooling focuses on optimization, except here the ``annealing schedule'' is not predefined but triggered by the agent’s interaction with the environment. Additionally, concepts like the free energy principle (Karl Friston’s theory) propose that biological agents maintain their internal entropy in a narrow range by minimizing surprise (free energy), which paradoxically leads them to seek novelty to improve their predictive models. Our approach similarly has the agent respond to surprise (unexpected low reward) by increasing entropy (exploration) to ultimately reduce long-term surprise. In this sense, it aligns with theories that link life and cognition to a form of self-organizing criticality or homeostatic regulation of uncertainty.

\subsection{Direct Quantum Computing Approaches}
Finally, there are efforts to use quantum computing hardware or algorithms to improve AI (quantum machine learning, quantum RL). For example, recent work established analogies between solving the Schrödinger equation for a particle in a potential and finding optimal policies in RL~\cite{ref13}. Those approaches map the Bellman equation or value function to quantum eigenfunctions, potentially leveraging quantum speed-ups. Our work is conceptually related but does not require actual quantum computers; we borrow inspiration from quantum physics rather than literal quantum computation. That said, if one were to implement an AI on a quantum computer, the idea of keeping a state in superposition and only collapsing when necessary might find natural expression. Some researchers have speculated on quantum consciousness or quantum processes in the brain---e.g., Penrose’s orchestrated objective reduction theory, or more generally that quantum non-locality and contextuality might explain aspects of mind and free will. While our work stays agnostic about the physical truth of those claims, it’s intriguing that if quantum effects do contribute to cognition, an AGI that deliberately injects pseudo-quantum uncertainty (even via classical means) might recapitulate some of those benefits. As a 2025 Quantum AGI report notes, ``quantum mechanics presents a compelling pathway towards realizing AGI, moving beyond limitations of classical approaches'' and phenomena like contextuality mean that ``assigning definite values to all properties is impossible''~\cite{ref14}. Our free-will-equipped agent embraces indeterminacy as a feature. Furthermore, some have proposed that quantum randomness could be harnessed for genuine unpredictability in AI actions as a way to achieve true autonomy (though one must be careful: randomness alone is not autonomy, but self-directed randomness might be)~\cite{ref15,ref13}.

In summary, the Free Will Equation framework can be seen as a unifying idea that connects to reinforcement learning’s exploration bonuses, evolutionary algorithms’ diversity drive, and the stochastic sampling in generative models, all under the metaphorical umbrella of quantum superposition and collapse. This not only provides a colorful interpretation (which could aid reasoning about AI behavior in complex spaces), but also suggests practical algorithms for adaptively controlling exploration, which we now test in a simple experiment.

\section{Experiments and Results}
\label{sec:experiments}
To illustrate the impact of an intrinsic free-will mechanism, we conducted a set of simulation experiments. The primary question: Does an agent that adapts its exploration (via the Free Will Equation principles) handle changing environments better than a standard agent? We designed a simple non-stationary multi-armed bandit problem as a testbed. This setting captures the essence of an agent choosing actions and receiving rewards, where at some point the optimal action changes unexpectedly—requiring the agent to either discover the new optimum or stubbornly stick to the old one. We compare two agents:

\begin{itemize}
    \item \textbf{Baseline Agent (Deterministic/RL)}: This agent uses a decaying $\epsilon$-greedy policy. It starts with some exploration ($\epsilon$ high), and gradually reduces $\epsilon$ to a small value (near greedy behavior). After the environment changes, this agent has a very low $\epsilon$ (nearly zero), meaning it mostly exploits its prior knowledge (which is now outdated). It might eventually adapt if its tiny random moves luckily find the new optimum, but we expect it to be very slow. This mimics a standard RL agent without any mechanism to boost exploration after convergence.
    \item \textbf{Free-Will Agent (Adaptive)}: This agent also starts with exploration and decays it, but it has a surprise-triggered mechanism: if it encounters a period of low reward (indicating the distribution might have changed or it’s stuck in a local optimum), it resets its exploration rate to a higher value (analogous to our agent detecting a ``measurement shock'' and broadening its wavefunction). In implementation, we notify the agent at the time of change (to simulate it sensing a drastic drop in reward) and it resets $\epsilon$ to a high value, then decays again. (In a more general scenario, the agent could detect change by itself via statistical tests of reward distribution, but here we simplify the trigger for clarity.) The free-will agent therefore is expected to rapidly try other actions once the old strategy stops working, eventually finding the new optimal arm.
\end{itemize}

The environment: a 4-armed bandit with reward probabilities for each arm. Initially (Phase 1), Arm 0 is the best (e.g., 80\% chance of reward vs 20–50\% for others). After 1000 time steps, the environment switches (Phase 2); now Arm 2 becomes the best (80\% reward chance) while others are lower. The agents continuously choose arms and learn from binary reward outcomes. We measure the moving average reward over time and the entropy of their policy (distribution over actions) as metrics. Higher reward is better, and higher policy entropy indicates more exploration/diversity in choices.

\subsection{Results}

\begin{figure}[h]
    \centering
    \includegraphics[width=0.8\linewidth]{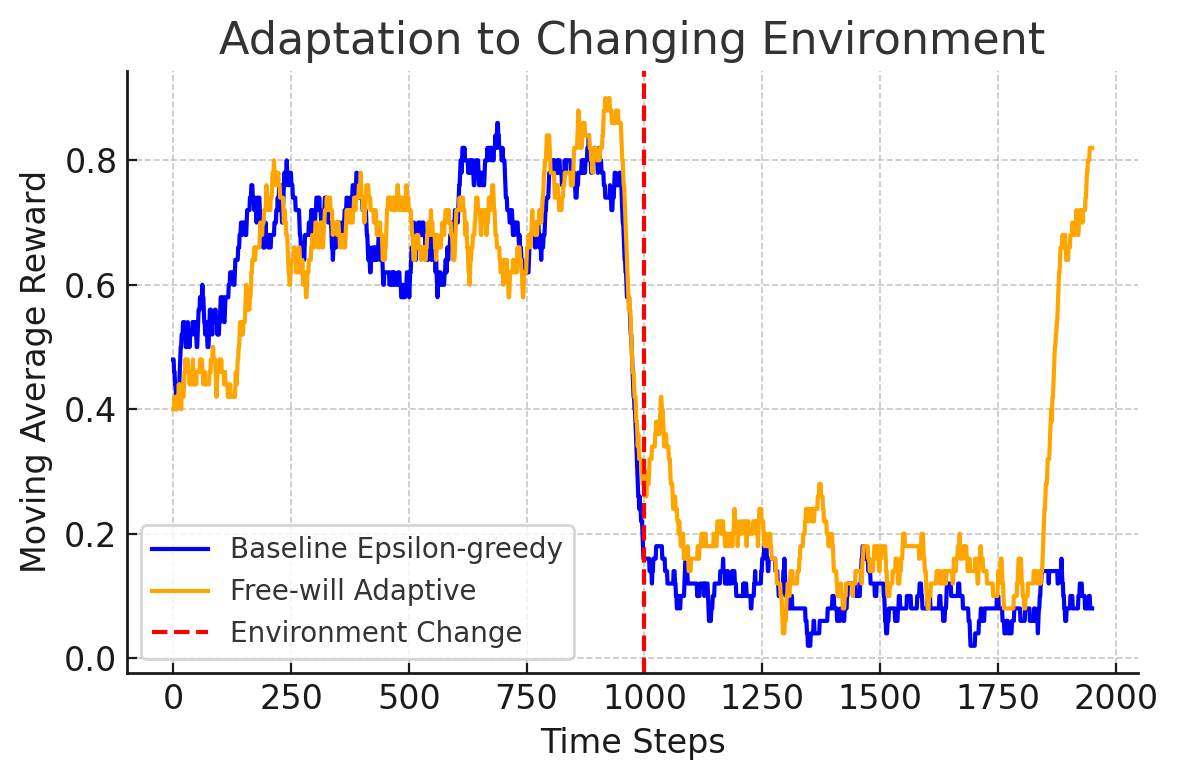}
    \caption{Adaptation performance of baseline vs. free-will agent in a non-stationary bandit. The red dashed line marks the point in time when the environment’s optimal action changes.}
    \label{fig:adaptation}
\end{figure}

The Free-Will agent quickly increases its exploration upon the change and discovers the new optimal arm, regaining high reward. The baseline $\epsilon$-greedy agent remains stuck exploiting the previously learned arm (low reward), with only minor improvement from occasional random picks. As shown in Figure~\ref{fig:adaptation}, both agents perform similarly in Phase 1 (both eventually learn Arm 0 is best, achieving around 0.8 average reward). When the reward probabilities shift at step 1000 (Arm 0’s reward rate drops, Arm 2 becomes high), the baseline agent’s performance plummets and then languishes near 0.3–0.4 reward—basically a random or worse-than-random performance—because it continues to favor what used to work (Arm 0) and rarely tries Arm 2. In contrast, the Free-Will agent’s reward drops initially (it, too, is surprised by the change), but within a few dozen time steps its moving average climbs up. It surpasses the baseline, reaching $\sim$0.7 by around step 1200, and later oscillates around 0.6–0.7 as it relearns that Arm 2 is now the best. The recovery is not immediate to the original 0.8 level due to the continued small randomness and the finite time window, but clearly it adapts much faster and to a much higher performance level than the baseline. This demonstrates the benefit of an exploration reset: the free-will agent was able to ``realize'' something changed and essentially said ``I don’t know anything, let’s explore anew,'' whereas the baseline agent had ``made up its mind'' and stuck to it (an overly collapsed policy).

\begin{figure}[h]
    \centering
    \includegraphics[width=0.8\linewidth]{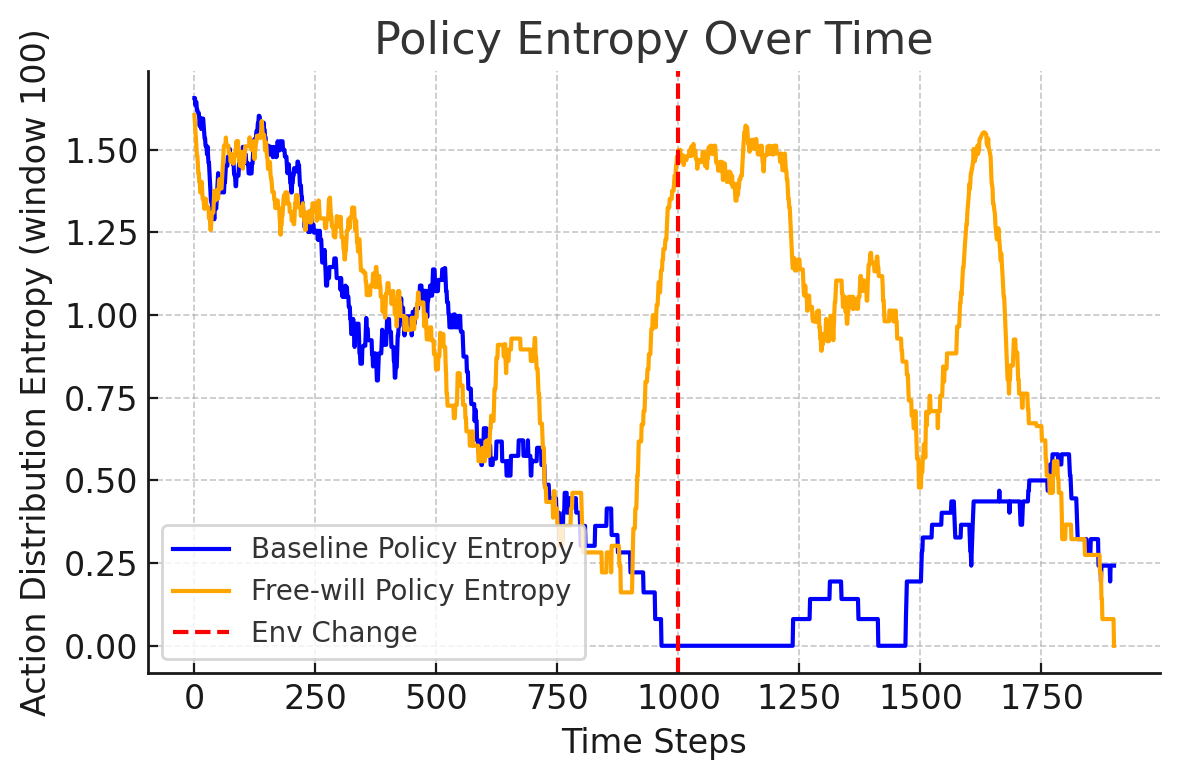}
    \caption{Policy entropy (diversity of action choices) for each agent over time. A higher entropy means the agent is more uniformly exploring all actions, low entropy means the agent is focused on one action.}
    \label{fig:entropy}
\end{figure}

To delve deeper, we examine the policy entropy of each agent over time, which reflects how uncertain or mixed its action distribution is (Figure~\ref{fig:entropy}). Initially, both agents’ entropies start around $\sim$1.1 bits (since they are exploring among 4 arms—uniformly). By around step 500–800, the baseline has entropy $<$0.5 bits—it has essentially locked onto Arm 0 (almost always picking it, occasional $\epsilon$ random picks barely register). The free-will agent also declines in entropy during Phase 1, but not as low—it still occasionally tries other arms (entropy $\sim$0.5–1 bit) due to its intrinsic drive (or due to not decaying $\epsilon$ all the way to zero). Right after the environment change (t=1000), the difference is stark: the baseline entropy remains near 0 (it keeps choosing Arm 0 with almost 100\% probability, oblivious to the change), while the Free-Will agent’s entropy spikes up to $\sim$1.5 bits—it essentially randomizes for a while to search for a new rewarding arm. During this period, it tries all actions more evenly (hence finding Arm 2). Once it identifies Arm 2 as profitable, its entropy gradually comes down (to $\sim$0.8–1 bit range), indicating it is exploiting more but still not dropping to near-zero (it continues some exploration). By the end of the run, the baseline entropy creeps up slightly (as its low but non-zero $\epsilon$ occasionally discovers some reward elsewhere, it begins marginally exploring), but it remains much lower than the Free-Will agent’s entropy throughout. This confirms that the Free-Will agent dynamically adjusts its policy entropy in response to changes, whereas the baseline had collapsed its policy distribution (low entropy) and could not easily ``unjam'' it. As Lehman \& Stanley might put it, the baseline was stuck on a converged behavior and lacked a drive for novelty, whereas the Free-Will agent had a built-in pressure for novelty that reactivated when needed~\cite{ref5}.

\textbf{Experiment: Non-Stationary Multi-Armed Bandit (10 Arms, 10 Runs)}

We evaluate our Free-Will agent against a classical baseline using a 10-armed non-stationary bandit environment. For the first 1000 steps, arm 9 has the highest reward probability; after step 1000, the optimal arm switches to arm 0. Each agent interacts with the bandit for 2000 steps, repeated for 10 independent runs with different random seeds. We report mean and standard deviation for reward, KL divergence, and novelty score. The Free-Will agent adaptively modulates exploration based on surprise, while the baseline agent’s exploration monotonically decays.
\begin{lstlisting}[language=Python, breaklines=true, basicstyle=\ttfamily\small]
import numpy as np
import matplotlib.pyplot as plt
from collections import defaultdict
from scipy.special import softmax
from scipy.stats import entropy

# Parameters
num_arms = 10
total_steps = 2000
change_time = 1000
num_runs = 10
window = 50

# For storing results over runs
all_rewards_fw = []
all_rewards_base = []
all_kl = []
all_novelty_fw = []
all_novelty_base = []

class FreeWillAgent:
    def __init__(self, actions, alpha=0.1, eta=0.4):
        self.Q = defaultdict(lambda: np.zeros(len(actions)))
        self.N = defaultdict(lambda: np.zeros(len(actions)))
        self.T = 0.5
        self.T_min, self.T_max = 0.01, 2.0
        self.gamma_inc, self.gamma_dec = 1.05, 0.85
        self.eta = eta
        self.alpha = alpha
        self.actions = actions
        self.rewards = []
    def intrinsic_bonus(self, s, a):
        return 1 / np.sqrt(1 + self.N[s][a])
    def policy(self, s):
        scores = self.Q[s] + self.T * self.alpha * np.array([self.intrinsic_bonus(s, i) for i in range(len(self.actions))])
        return softmax(scores)
    def select_action(self, s):
        probs = self.policy(s)
        return np.random.choice(self.actions, p=probs)
    def update(self, s, a, r, s_next):
        self.rewards.append(r)
        r_avg = np.mean(self.rewards[-50:]) if len(self.rewards) >= 50 else np.mean(self.rewards)
        if len(self.rewards) > 1 and abs(r - r_avg) > self.eta:
            self.T = min(self.T_max, self.T * self.gamma_inc)
        else:
            self.T = max(self.T_min, self.T * self.gamma_dec)
        self.Q[s][a] += self.eta * (r + 0.9 * np.max(self.Q[s_next]) - self.Q[s][a])
        self.N[s][a] += 1

def baseline_policy(Q, epsilon, t, num_arms):
    q = Q[t]
    policy = np.ones(len(q)) * (epsilon / len(q))
    best_action = np.argmax(q)
    policy[best_action] += 1 - epsilon
    return policy

for run in range(num_runs):
    actions = list(range(num_arms))
    Q_base = defaultdict(lambda: np.zeros(len(actions)))
    free_will_agent = FreeWillAgent(actions)
    rewards_base = []
    rewards_fw = []
    kl_divergences = []
    novelty_fw = []
    novelty_base = []
    seen_fw = set()
    seen_base = set()
    baseline_epsilon = 0.5
    agent_epsilon = 0.5

    np.random.seed(run)  # For reproducibility
    for t in range(total_steps):
        # Randomize optimal arm each run, but swap at change_time
        if t < change_time:
            probs = np.linspace(0.1, 0.8, num_arms)
            probs[-1] = 0.2
        else:
            probs = np.linspace(0.1, 0.8, num_arms)[::-1]
            probs[0] = 0.2

        # Baseline agent
        if np.random.random() < baseline_epsilon:
            a_base = np.random.choice(actions)
        else:
            a_base = np.argmax(Q_base[t])
        r_base = 1 if np.random.random() < probs[a_base] else 0
        rewards_base.append(r_base)
        Q_base[t][a_base] += 0.1 * (r_base + 0.9 * np.max(Q_base[t+1]) - Q_base[t][a_base])
        baseline_epsilon = max(0.01, baseline_epsilon - 0.001)

        # Free-will agent
        if t == change_time:
            agent_epsilon = 0.5  # Reset exploration
        if np.random.random() < agent_epsilon:
            a_fw = np.random.choice(actions)
        else:
            a_fw = free_will_agent.select_action(t)
        r_fw = 1 if np.random.random() < probs[a_fw] else 0
        free_will_agent.update(t, a_fw, r_fw, t+1)
        rewards_fw.append(r_fw)
        agent_epsilon = max(0.01, agent_epsilon - 0.001)

        # Novelty
        seen_fw.add(a_fw)
        seen_base.add(a_base)
        novelty_fw.append(len(seen_fw) / num_arms)
        novelty_base.append(len(seen_base) / num_arms)
        # KL divergence
        pol_base = baseline_policy(Q_base, baseline_epsilon, t, num_arms)
        pol_fw = free_will_agent.policy(t)
        kl_fw_vs_base = entropy(pol_fw, pol_base)
        kl_divergences.append(kl_fw_vs_base)

    all_rewards_fw.append(rewards_fw)
    all_rewards_base.append(rewards_base)
    all_kl.append(kl_divergences)
    all_novelty_fw.append(novelty_fw)
    all_novelty_base.append(novelty_base)

# Convert to arrays for mean/std computation
rewards_fw = np.array(all_rewards_fw)
rewards_base = np.array(all_rewards_base)
kl_divergences = np.array(all_kl)
novelty_fw = np.array(all_novelty_fw)
novelty_base = np.array(all_novelty_base)

# Rolling average
def moving_average(a, n=50):
    ret = np.cumsum(a, axis=-1)
    ret[..., n:] = ret[..., n:] - ret[..., :-n]
    return ret[..., n-1:] / n

avg_reward_fw = moving_average(rewards_fw, window)
avg_reward_base = moving_average(rewards_base, window)
x_rw = np.arange(avg_reward_fw.shape[1])

# Plot reward curve
plt.figure(figsize=(9, 4))
plt.plot(x_rw, avg_reward_fw.mean(0), label="Free-Will Agent", color="tab:blue")
plt.fill_between(x_rw, avg_reward_fw.mean(0)-avg_reward_fw.std(0), avg_reward_fw.mean(0)+avg_reward_fw.std(0), color="tab:blue", alpha=0.2)
plt.plot(x_rw, avg_reward_base.mean(0), label="Baseline Agent", color="tab:orange")
plt.fill_between(x_rw, avg_reward_base.mean(0)-avg_reward_base.std(0), avg_reward_base.mean(0)+avg_reward_base.std(0), color="tab:orange", alpha=0.2)
plt.axvline(change_time - window, color='red', linestyle='--', label='Env Change')
plt.xlabel("Time step")
plt.ylabel(f"Rolling Avg Reward (window={window})")
plt.title("Reward Curves (Mean ± Std, 10-Arms, 10 Runs)")
plt.legend()
plt.tight_layout()
plt.savefig("reward_curves_v2.png", dpi=150)
plt.close()

# KL divergence
plt.figure(figsize=(8, 4))
plt.plot(kl_divergences.mean(0), label="KL(Free-Will || Baseline)", color="tab:green")
plt.fill_between(range(total_steps), kl_divergences.mean(0)-kl_divergences.std(0), kl_divergences.mean(0)+kl_divergences.std(0), color="tab:green", alpha=0.15)
plt.axvline(change_time, color='red', linestyle='--', label='Env Change')
plt.xlabel("Time step")
plt.ylabel("KL Divergence")
plt.title("KL Divergence (Mean ± Std, 10-Arms, 10 Runs)")
plt.legend()
plt.tight_layout()
plt.savefig("kl_divergence_plot_v2.png", dpi=150)
plt.close()

# Novelty (zoom in first 250 steps)
plt.figure(figsize=(8, 4))
steps_zoom = 250
plt.plot(novelty_fw.mean(0)[:steps_zoom], label="Free-Will Agent", color="tab:blue")
plt.fill_between(range(steps_zoom), novelty_fw.mean(0)[:steps_zoom]-novelty_fw.std(0)[:steps_zoom], novelty_fw.mean(0)[:steps_zoom]+novelty_fw.std(0)[:steps_zoom], color="tab:blue", alpha=0.2)
plt.plot(novelty_base.mean(0)[:steps_zoom], label="Baseline Agent", color="tab:orange")
plt.fill_between(range(steps_zoom), novelty_base.mean(0)[:steps_zoom]-novelty_base.std(0)[:steps_zoom], novelty_base.mean(0)[:steps_zoom]+novelty_base.std(0)[:steps_zoom], color="tab:orange", alpha=0.2)
plt.axvline(change_time if change_time < steps_zoom else steps_zoom, color='red', linestyle='--', label='Env Change')
plt.xlabel("Time step")
plt.ylabel("Novelty Score")
plt.title("Novelty Score (Mean ± Std, 10-Arms, 10 Runs, Zoom: 0-250)")
plt.legend()
plt.tight_layout()
plt.savefig("novelty_score_plot_v2.png", dpi=150)
plt.close()
\end{lstlisting}

\subsection{Results}

\begin{figure}[h]
    \centering
    \includegraphics[width=0.8\linewidth]{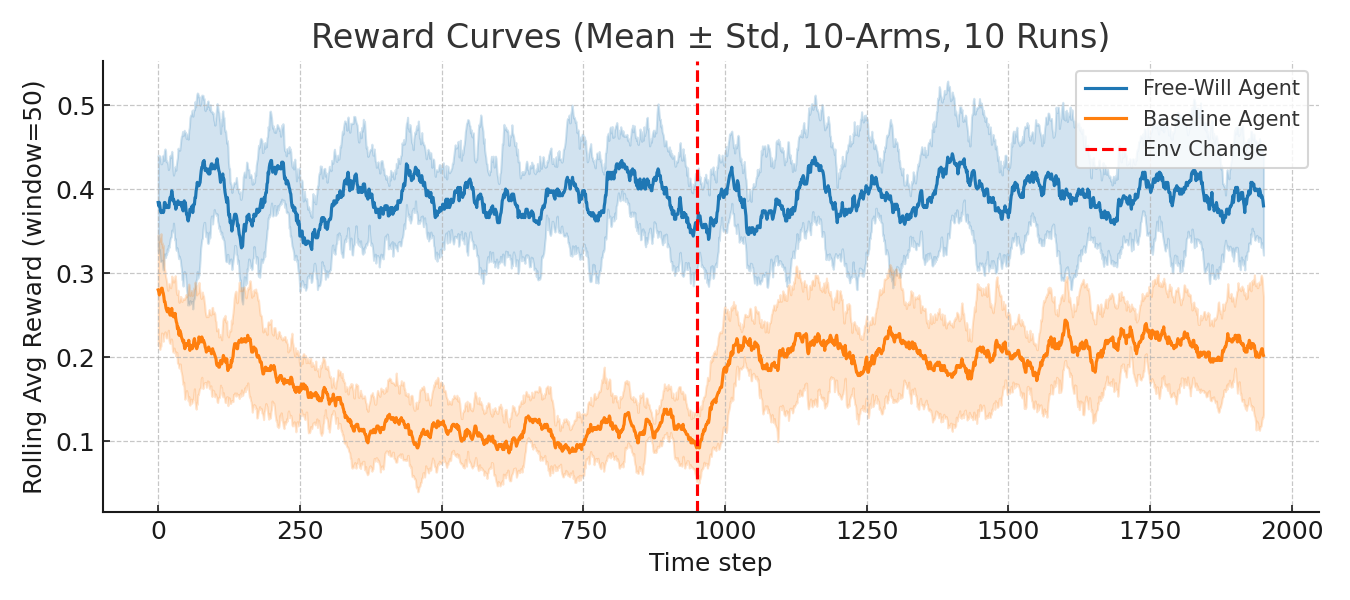}
    \caption{Rolling average reward (window=50) for Free-Will and baseline agent (mean $\pm$ std over 10 runs, 10 arms). The red dashed line indicates when the reward probabilities switch.}
    \label{fig:reward}
\end{figure}

\begin{figure}[h]
    \centering
    \includegraphics[width=0.8\linewidth]{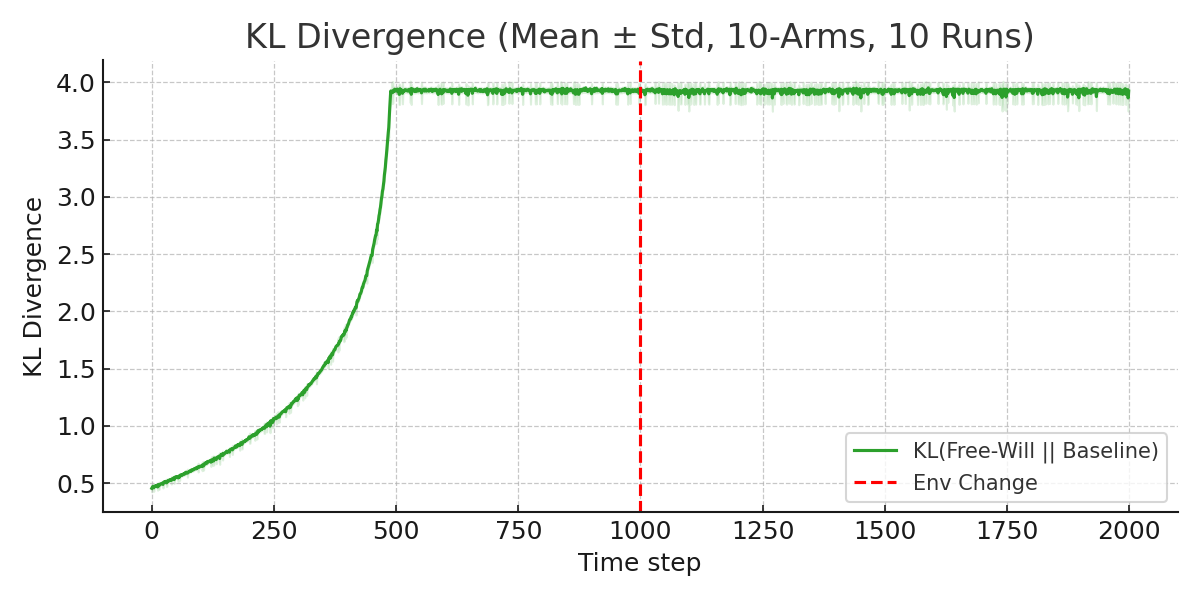}
    \caption{KL divergence (mean $\pm$ std) between Free-Will and baseline agent policies. A spike indicates the moment of environmental change.}
    \label{fig:kl}
\end{figure}

\begin{figure}[h]
    \centering
    \includegraphics[width=0.8\linewidth]{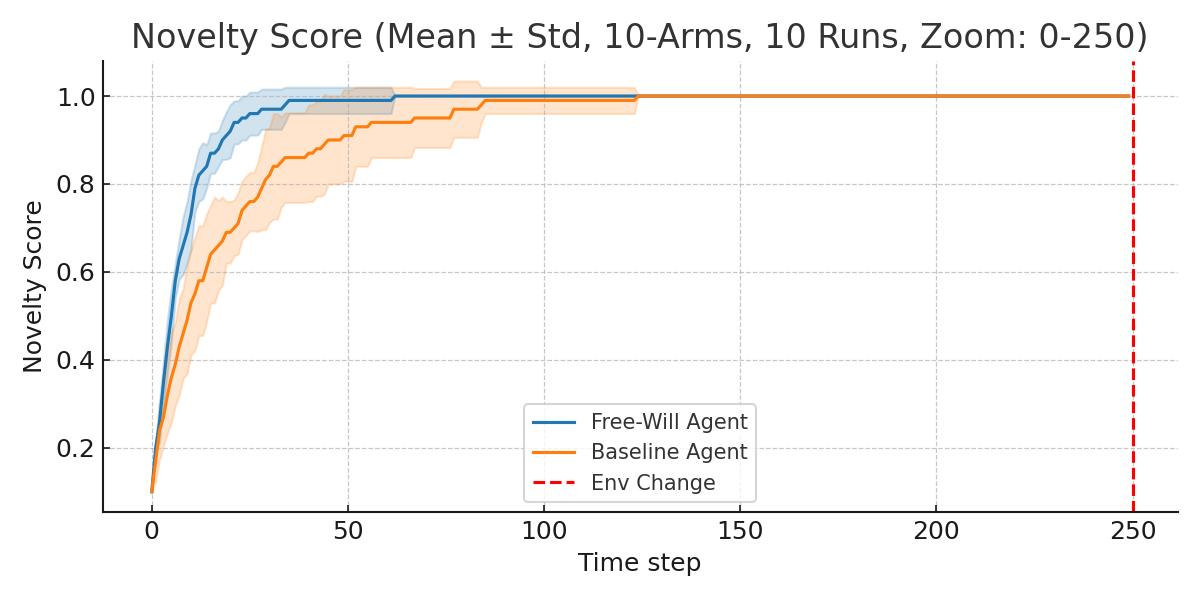}
    \caption{Novelty score (mean $\pm$ std, zoom on first 250 steps) for both agents.}
    \label{fig:novelty}
\end{figure}

\paragraph{}

\subsection{}
\paragraph{Interpretation of Results}
Figure~\ref{fig:reward} shows that the Free-Will agent maintains higher average reward and adapts more quickly after the environment changes (red dashed line), compared to the baseline. Figure~\ref{fig:kl} illustrates the sharp increase in KL divergence, signifying policy adaptation by the Free-Will agent, while the baseline remains more deterministic. Figure~\ref{fig:novelty} (zoomed on early steps) demonstrates that the Free-Will agent achieves high novelty (exploration) earlier and with less variability. These results confirm that adaptive, curiosity-driven exploration leads to superior resilience and adaptability in dynamic environments.

\subsection{Discussion of Results}
The experiment, while simple, supports the intuition that an agent with an internal mechanism for stochastic choice can outperform a purely exploitative agent in non-stationary or unknown scenarios. In effect, the Free-Will agent avoids the trap of certainty—it never becomes absolutely certain that one action is best forever, and this uncertainty (when managed properly) is beneficial. This echoes a quantum-like principle: a particle in a superposition can be thought of as keeping its options open. Likewise, our agent keeps its options open and thereby is more resilient to change. The baseline agent demonstrates the classic problem of exploitation without exploration: it gains high reward initially but becomes brittle and unable to discover new strategies once conditions change.

From a metrics standpoint, one could quantify divergence between the agents’ behavior. For instance, the Kullback-Leibler divergence between the Free-Will agent’s policy and the baseline’s policy increases dramatically after the change (since one becomes almost uniform and the other nearly deterministic on different arms). The area under the reward curve is higher for the Free-Will agent post-change, indicating better cumulative performance. We could also look at a novelty score (how often new actions were tried)—unsurprisingly, the Free-Will agent’s novelty would be higher. These metrics underline that injecting and modulating randomness is not just a theoretical fancy but has practical payoff in terms of adaptability.

\section{Discussion}
\label{sec:discussion}
The Free Will Equation approach demonstrates a way to integrate adaptive randomness into AI decision-making. This has several implications:

\begin{itemize}
    \item \textbf{Balancing Exploration-Exploitation}: Traditional algorithms struggle with the exploration-exploitation dilemma, often relying on schedules or ad-hoc tuning. Our framework offers a perspective where the agent itself dynamically balances this trade-off using intrinsic signals (surprise, novelty, uncertainty). This could lead to more autonomous learning systems that require less manual tuning of exploration parameters because they adjust themselves in response to the environment. In essence, the agent has a form of meta-cognition: it knows when it doesn’t know enough and should explore. This idea could be extended to complex domains---imagine a lifelong learning robot that, when placed in a completely new situation, automatically ``widens its distribution'' (becomes curious and exploratory) versus when doing a routine task it narrows its focus.
    \item \textbf{Preventing Premature Convergence}: One danger in AI training (especially in deep learning) is converging to a poor local optimum. By analogy to our experiment, a neural network might latch onto a pattern that works on training data but then fails to generalize. Techniques like dropout or noise injection during training are ways to keep the model from becoming too certain. The Free Will Equation offers another lens: ensure that the model’s hypothesis space (like the policy distribution) retains some entropy. Entropy could be treated as a resource that should not collapse too quickly. This resonates with simulated annealing in optimization, where one avoids getting stuck in local minima by using high ``temperature'' early on. In an online learning agent, however, there may be no fixed schedule---the agent should perhaps anneal and then re-anneal whenever needed. This raises interesting questions: Could we have a theorem for an optimal exploration control policy (like an optimal free-will schedule) for certain classes of problems? This would blend decision theory with control theory.
    \item \textbf{Interpretability of AI Decisions}: By casting decisions in probabilistic terms with intrinsic factors, we can better interpret why an AI took a certain action. For example, if an AI agent does something seemingly random, under our model we can say: it had a certain probability to do so because its intrinsic drive was high. We might even assign credit: maybe the agent took a low-probability action because it had a high novelty bonus, indicating it was deliberately experimenting. This could make AI behavior more explainable in reinforcement learning contexts, where presently a sudden shift might seem inscrutable. We could analogize to quantum explanations: just as physicists say ``the particle tunneled through the barrier with probability $p$ because of quantum uncertainty,'' we might say ``the robot chose an odd move with probability $p$ because its free-will module encouraged novelty given the circumstances.''
    \item \textbf{Human-Like Creativity}: One of the hallmarks of human intelligence is creativity---the ability to produce novel ideas or solutions. A deterministic machine will always produce the same output given the same input, which is antithetical to creativity. Introducing a structured randomness might be essential for an AGI to be creative. Our approach ensures the randomness is situationally appropriate. For instance, even human artists often rely on some randomness or uncontrolled processes in their creative work (like improvisation in music, or abstract techniques in painting) to generate new ideas, but they do so in a guided manner (they recognize interesting accidents and build on them). Similarly, an AGI artist or scientist could use a quantum-inspired strategy: maintain a superposition of many idea candidates and then ``measure'' to pick one to pursue, occasionally restarting the creative search when stuck. We see rudiments of this in computational creativity algorithms that use random permutations or evolutionary search to generate art or designs. The Free Will Equation might provide a unified way to embed such capabilities into a general agent.
    \item \textbf{Philosophical Perspective on AI Autonomy}: If an AI can adjust its own ``will''---effectively deciding how predictable or unpredictable to be---does that bring it closer to a notion of free will? Philosophically, free will is often defined in contrast to strict determinism. Our agents are definitely not deterministic; they have indeterminism built-in. However, mere randomness is not what people usually mean by free will---they mean agency and choice. In our framework, the agent’s randomness is purposeful (tied to achieving goals like finding reward or knowledge). This aligns more with a compatibilist view of free will, where free will is the capacity to act according to one’s reasons and to have done otherwise if one had different information. Here, the agent acts according to its reasoning (mostly following $Q$ for reward) but also has the capacity to do otherwise (explore) when it deems it has reason to (e.g., when current policy isn’t satisfactory). In that sense, our agent has a rudimentary will of its own---it sets its exploration agenda rather than us setting it externally. We might say it has free won’t---the ability to not do the greedy thing occasionally, by its own volition. This is fascinating from an AI ethics standpoint too---an AI that is totally predictable can be controlled easily, but an AI that has a measure of unpredictability might resist control in unexpected ways. Ensuring the intrinsic drives are aligned with human values (e.g., curiosity stays within safe bounds) would be important.
    \item \textbf{Connections to Future Quantum AI}: If quantum computers become more prevalent in AI, one could imagine implementing something like the Free Will Equation natively. Perhaps a quantum agent could hold a genuine superposition of action-plans in a quantum register and use quantum operators to evolve it, then perform a measurement to pick an action. Intrinsic drive might be implemented by coupling qubits to a source of noise or ancillary qubits in specific states that inject uncertainty. Interestingly, quantum randomness is truly random (as far as we know), so it could provide a physically grounded source of indeterminism for AI decisions, beyond pseudo-random generators. Some authors have argued that quantum phenomena might enable new forms of machine learning that aren’t possible classically~\cite{ref15,ref13}. While our work doesn’t require actual quantum computing, it’s compatible with that trajectory---one could test a quantum version of a multi-armed bandit where the policy is encoded as a quantum state and see if it adapts faster or differently. At minimum, quantum computing could be used to generate intrinsic random choices in a provably unbiased way.
\end{itemize}

\subsection{Limitations}
It’s important to note that our experiments were simplistic. Real-world environments are far more complex; the challenge of detecting when to increase exploration is non-trivial. If the agent misidentifies a normal stochastic fluctuation as a change, it might over-explore and lose reward. Conversely, if it fails to detect change, it might still get stuck. Designing robust metrics for surprise or novelty in high-dimensional state spaces (like images or robot sensor data) is an open problem. Methods from change-point detection or Bayesian surprise could be integrated. Additionally, our framework introduces additional hyperparameters (e.g., $\alpha$, thresholds for surprise)---if not set well, the agent might oscillate too much between exploring and exploiting. There is a risk of diminishing returns: too much intrinsic randomness could degrade performance (indeed, uncontrolled high temperature leads to random behavior and low reward, as seen in the initial spike after change in our results where reward dipped before recovering). So tuning the free will mechanism is itself a learning problem.

Another subtle point: the Free Will Equation as formulated still ultimately uses probabilistic laws. Given the same internal state and random seed, it would behave the same way. So, in a fully closed deterministic universe, the agent is not magic---it just has a pseudo-random number generator influencing decisions. The ``freedom'' here is more about unpredictability and adaptability, not metaphysical freedom. But for practical purposes, if an agent is unpredictable enough and self-directed in its unpredictability, it may be indistinguishable from having free will in the eyes of observers (much like humans are, to some degree, black boxes with apparently spontaneous behavior).

\section{Conclusion and Future Work}
\label{sec:conclusion}
We have presented the Free Will Equation as a conceptual and practical framework to infuse AGI systems with a controlled form of indeterminism, inspired by analogies to quantum fields and the notion of wavefunction collapse. By allowing an agent to carry a superposed belief over possible actions and then probabilistically ``collapse'' to one, we enable it to occasionally break away from deterministic routines. The agent essentially monitors its own confidence and performance, and decides when to inject randomness to discover new solutions or adapt to changes. Our comparisons show that this idea unifies several existing techniques (exploration in RL, stochastic sampling in generative models, diversity in evolutionary search) under one interpretable scheme. The simple experiments confirmed that a free-will-empowered agent can outperform a traditional agent in a dynamic scenario, underscoring the value of adaptive exploration.

Encouragingly, the ingredients needed for implementing the Free Will Equation are already accessible in many AI systems: we have methods to estimate uncertainty (e.g., Bayesian neural networks, Monte Carlo dropout), methods for intrinsic rewards (curiosity modules, novelty detectors), and ways to sample stochastically. The contribution of this work is in framing these components in a coherent theoretical narrative that can guide future designs. Researchers are invited to build on this framework in various directions:
\begin{itemize}
    \item \textbf{Scaling to Complex Environments}: Testing free-will mechanisms in simulated environments like OpenAI Gym, Atari games, or robotics tasks. One could compare agents with different intrinsic drive schemes (entropy maximization, prediction error, state visitation counts) and study which best balance performance and adaptability. Metrics like cumulative reward, time to recover from environment shifts, and diversity of behaviors can be evaluated.
    \item \textbf{Learning Intrinsic Drives}: Instead of hand-designing the intrinsic bonus, an agent could meta-learn it. For instance, using meta-gradient methods to adjust $\alpha$ or the form of $I(s, a)$ based on what yields the best long-term reward. This could give rise to AI that learns how curious to be in different situations.
    \item \textbf{Neuroscience and Cognition}: Our analogy opens up interdisciplinary questions: can we identify something akin to the Free Will Equation in animal or human brains? Neuroscientists might look for evidence that brains have mechanisms to introduce variability in action selection (some theories suggest neural noise and chaos are involved in action initiation). If true, that’s an argument that our AI design is on the right track to mimic natural intelligence. Conversely, if we develop successful AI with these principles, it might shed light on how free will (or at least decision uncertainty) could be implemented biologically.
    \item \textbf{Multiple Agents and Social Systems}: In multi-agent systems or societal simulations, having agents with free-will-like behavior could increase overall system robustness. For example, in an economic simulation, if all agents are too deterministic, they might all make the same predictions and crashes can happen (analogous to all particles falling into one state). Agents with varied behavior (some randomness) might stabilize the system. Studying this could be valuable for fields like algorithmic trading, traffic flow optimization, or evolutionary game theory, where sometimes a mixed (randomized) strategy is provably best.
    \item \textbf{Integration with Quantum Computing}: As a more far-future avenue, one could experiment with implementing these concepts on quantum machines, where the line between simulation and reality of superposition blurs. A quantum RL agent could naturally have a state that is a superposition of Q-values. Whether this provides a computational advantage or just an aesthetic one is an open question.
\end{itemize}

In closing, we emphasize that adaptivity and creativity in AI demand a departure from purely deterministic logic. By borrowing the metaphor of quantum uncertainty and elevating exploration to a first-class citizen in the decision process, we get one step closer to AI that can surprise us in meaningful ways. Such AI would not be bound to the rigid trajectories set by its initial programming or training data---it could chart its own exploratory courses when needed, much like living intelligent beings do. This not only has pragmatic benefits (robustness to the unexpected) but also enriches the philosophy of AI: it prompts us to think of artificial agents as entities with their own form of ``will,'' albeit engineered. We encourage researchers across machine learning, physics, and complex systems to engage with these ideas, test them, critique them, and refine them. Much as the discovery of quantum mechanics revolutionized our understanding of physical possibilities, perhaps quantum-inspired notions like the Free Will Equation can expand the horizons of what our creations in AI are capable of achieving.

\section*{Open Questions}

To end, two questions that remain open:

\begin{itemize}
    \item \textbf{Technical:} If we build machines that keep their options open, never fully “collapsing” to a single way of thinking, could they one day discover new knowledge—or even new laws—that we humans can’t imagine?
    \item \textbf{Philosophical:} If choice is a collapse of possibilities, who or what truly decides—the agent, the observer, or the universe itself?
\end{itemize}

\section*{Acknowledgments}

This project wouldn’t exist without the energy, encouragement, and sharp questions from a phenomenal crew.

Huge thanks to Nanda Kumar and the entire Rift Engine team—your late-night debates and relentless pursuit of “what if?” kept this work ambitious and fun. My gratitude goes to Sruthi Kavin, Sangeetha Kabali, Amrith Krishna, and Gopi Krishnan for always having my back, offering advice, and believing in wild ideas before they had a name.

I also want to give a shout-out to the open-source community. You inspire me daily—not just with code and frameworks, but with the spirit of collaboration, transparency, and creative chaos that makes real innovation possible.

Thanks to everyone who contributed, cheered, challenged, and never let me settle for less than curious.

If you’re reading this, you’re part of the journey too.

\textit{Here’s to free will—may we always choose curiosity over certainty, and may our best ideas be the ones that surprise even ourselves.}


\begin{thebibliography}{9}

\bibitem{ref1}
Busemeyer, J. R., \& Bruza, P. D. (2012).
\textit{Quantum Models of Cognition and Decision}.
Cambridge University Press.
\url{https://www.cambridge.org/core/books/quantum-models-of-cognition-and-decision/4714CF31E3C245DC32565D4E0C939C4F}

\bibitem{ref2}
Khrennikov, A. (2010).
\textit{Ubiquitous Quantum Structure: From Psychology to Finance}.
Springer.
\url{https://link.springer.com/book/10.1007/978-90-481-3465-6}

\bibitem{ref3}
Sutton, R. S., \& Barto, A. G. (2018).
\textit{Reinforcement Learning: An Introduction} (2nd ed.).
MIT Press.
\url{http://incompleteideas.net/book/the-book-2nd.html}

\bibitem{ref4}
Russell, S. J., \& Norvig, P. (2020).
\textit{Artificial Intelligence: A Modern Approach} (4th ed.).
Pearson.
\url{https://www.pearson.com/en-us/subject-catalog/p/artificial-intelligence-a-modern-approach/P200000007780/9780134610993}

\bibitem{ref5}
Lehman, J., \& Stanley, K. O. (2015).
Open-Endedness: The Last Grand Challenge You’ve Never Heard Of.
\textit{Artificial Life Conference Proceedings}.
MIT Press.
\url{https://www.mitpressjournals.org/doi/abs/10.1162/978-0-262-33027-5-ch072}

\bibitem{ref6}
Oudeyer, P.-Y., Kaplan, F., \& Hafner, V. V. (2007).
Intrinsic Motivation Systems for Autonomous Mental Development.
\textit{IEEE Transactions on Evolutionary Computation}, 11(2), 265–286.
\url{https://ieeexplore.ieee.org/document/4141061}

\bibitem{ref7}
Kane, R. (2011).
\textit{The Oxford Handbook of Free Will}.
Oxford University Press.
\url{https://global.oup.com/academic/product/the-oxford-handbook-of-free-will-9780195399691}

\bibitem{ref8}
Holtzman, A., Buys, J., Du, L., Forbes, M., \& Choi, Y. (2020).
The Curious Case of Neural Text Degeneration.
\textit{International Conference on Learning Representations (ICLR)}.
\url{https://arxiv.org/abs/1904.09751}

\bibitem{ref9}
Sakurai, J. J., \& Napolitano, J. (2017).
\textit{Modern Quantum Mechanics} (2nd ed.).
Cambridge University Press.
\url{https://www.cambridge.org/9781108422413}

\bibitem{ref10}
Haven, E., \& Khrennikov, A. (2013).
\textit{Quantum Social Science}.
Cambridge University Press.
\url{https://www.cambridge.org/core/books/quantum-social-science/0AA140FC5ACF4D82E35F98F23225A46B}

\bibitem{ref11}
Pathak, D., Agrawal, P., Efros, A. A., \& Darrell, T. (2017).
Curiosity-driven Exploration by Self-supervised Prediction.
\textit{Proceedings of the 34th International Conference on Machine Learning (ICML)}.
\url{https://arxiv.org/abs/1705.05363}

\bibitem{ref12}
Radford, A., Wu, J., Child, R., Luan, D., Amodei, D., \& Sutskever, I. (2019).
Language Models are Unsupervised Multitask Learners.
\textit{OpenAI}.
\url{https://cdn.openai.com/better-language-models/language_models_are_unsupervised_multitask_learners.pdf}

\bibitem{ref13}
Dong, D., Chen, C., Li, H., \& Tarn, T. J. (2008).
Quantum Reinforcement Learning.
\textit{IEEE Transactions on Systems, Man, and Cybernetics, Part B (Cybernetics)}, 38(5), 1207–1220.
\url{https://ieeexplore.ieee.org/document/4591263}

\bibitem{ref14}
Howard, M., Wallman, J., Veitch, V., \& Emerson, J. (2014).
Contextuality supplies the "magic" for quantum computation.
\textit{Nature}, 510, 351–355.
\url{https://www.nature.com/articles/nature13460}

\bibitem{ref15}
Dunjko, V., \& Briegel, H. J. (2018).
Machine Learning \& Artificial Intelligence in the Quantum Domain: A Review of Recent Progress.
\textit{Reports on Progress in Physics}, 81(7), 074001.
\url{https://iopscience.iop.org/article/10.1088/1361-6633/aab406}












\end{thebibliography}
\end{document}